\documentclass[10pt,twocolumn,letterpaper]{article}

\usepackage{cvpr}              % To produce the CAMERA-READY version
\usepackage{booktabs}
\usepackage{multirow}
\usepackage{siunitx}
\usepackage{graphicx}
\usepackage[table]{xcolor}
\definecolor{cvprblue}{rgb}{0.21,0.49,0.74}
\usepackage[pagebackref,breaklinks,colorlinks,allcolors=cvprblue]{hyperref}

\def\paperID{13433} % *** Enter the Paper ID here
\def\confName{CVPR}
\def\confYear{2026}

\newcommand\model{UniX}

\title{\model: Unifying Autoregression and Diffusion for Chest X-Ray \\ Understanding and Generation}

\author{
    Ruiheng Zhang$^{1,*}$, Jingfeng Yao$^{2,*}$, Huangxuan Zhao$^{1,*,\dagger}$, Hao Yan$^{1}$, Xiao He$^{1}$, 
    Lei Chen$^{2}$, \\
    Zhou Wei$^{1}$, Yong Luo$^{1}$, Zengmao Wang$^{1}$, Lefei Zhang$^{1}$, Dacheng Tao$^{3}$, Bo Du$^{1,\dagger}$ \\[0.5em]
    $^{1}$Wuhan University \\
    $^{2}$Huazhong University of Science and Technology \\
    $^{3}$Nanyang Technological University
}

\begin{document}
\maketitle
\begin{abstract}
Despite recent progress, medical foundation models still struggle to unify visual understanding and generation, as these tasks have inherently conflicting goals: semantic abstraction versus pixel-level reconstruction. Existing approaches, typically based on parameter-shared autoregressive architectures, frequently lead to compromised performance in one or both tasks.
To address this, we present \textbf{\model{}, a next-generation unified medical foundation model} for chest X-ray understanding and generation. \model{} decouples the two tasks into an autoregressive branch for understanding and a diffusion branch for high-fidelity generation. 
Crucially, a cross-modal self-attention mechanism is introduced to dynamically guide the generation process with understanding features. Coupled with a rigorous data cleaning pipeline and a multi-stage training strategy, this architecture enables synergistic collaboration between tasks while leveraging the strengths of diffusion models for superior generation.
On two representative benchmarks, \model{} achieves a \textbf{46.1\% improvement in understanding} performance (Micro-F1) and a \textbf{24.2\% gain in generation} quality (FD-RadDino), using only a quarter of the parameters of LLM-CXR. By achieving performance on par with task-specific models, our work establishes a scalable paradigm for synergistic medical image understanding and generation.
Codes and models are available at \href{https://github.com/ZrH42/UniX}{https://github.com/ZrH42/UniX}.
\end{abstract}

\begin{figure}[t]
  \centering
  \includegraphics[width=0.95\linewidth]{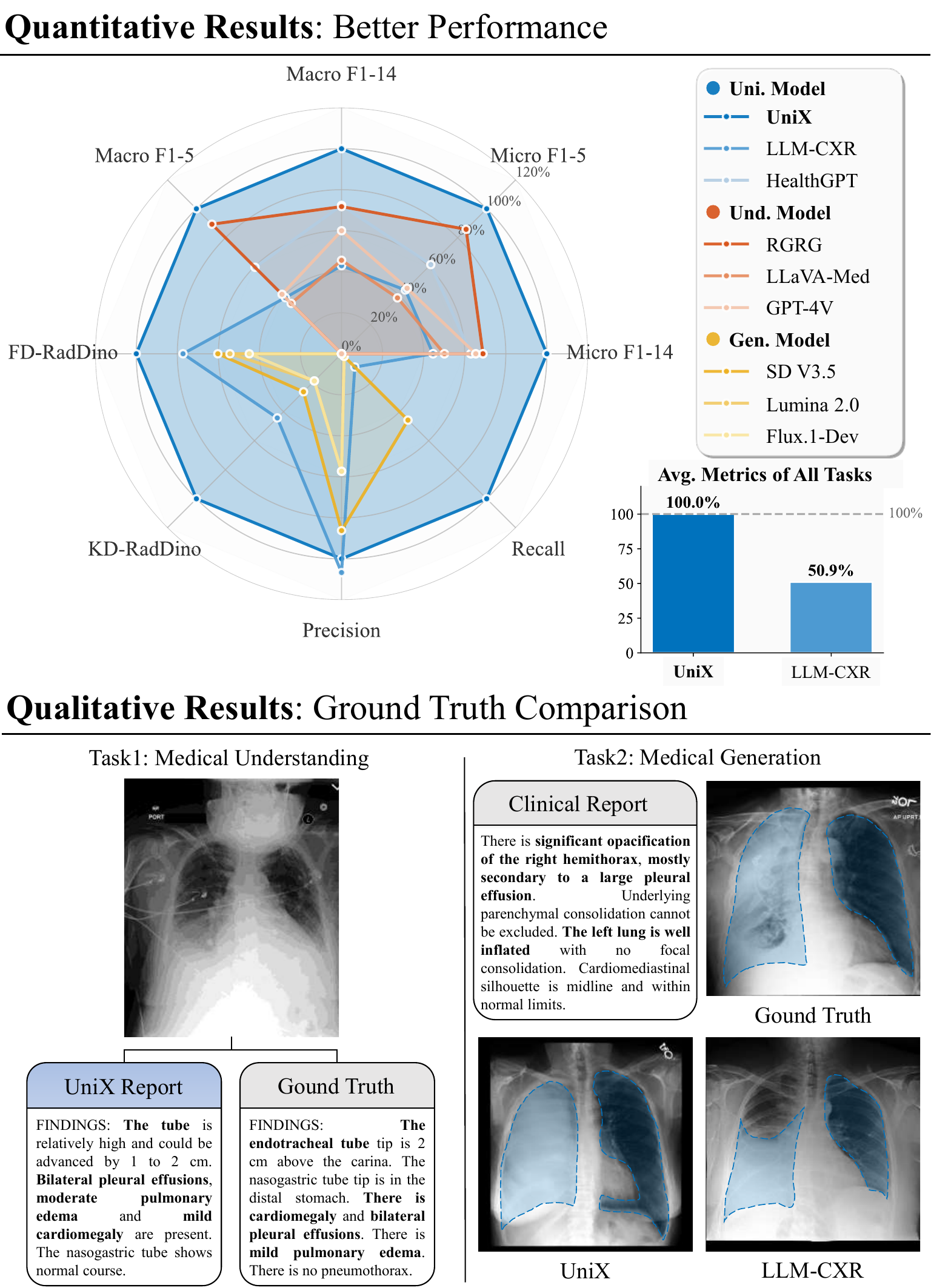}
  \caption{\textbf{The quantitative and qualitative results of \model}. Quantitative results show \model{}'s superiority over existing unified and single-task medical foundation models in understanding and generation. Qualitatively, \model{} enables multi-disease X-ray interpretation and high-fidelity medical image generation.}
  \label{fig:TopPic}
\end{figure}

\section{Introduction}
In recent years, vision–language pretraining–based medical foundation models have shown remarkable success in both understanding~\cite{yao2024eva, cui2024lkcell, ma2025fully, tu2024towards, li2023llava, chaves2024towards, tanno2025collaboration, bai2025bench} and generation~\cite{chambon2022roentgen, dutt2025chexgenbench, bluethgen2025vision, xu2024most, perez2024radedit,xu2025garamost} tasks. As research progresses, medical image understanding and generation are increasingly seen as interconnected tasks, where semantic reasoning and visual synthesis can mutually reinforce each other. This insight has spurred the development of unified medical foundation models~\cite{lee2023llm, zhang2025medunifier, kim2023unixgen, lin2025healthgpt} that aim to integrate both capabilities within a single framework.

However, unified modeling of these two capabilities is inherently challenging, given their fundamentally different objectives of semantic abstraction versus pixel-level reconstruction. Existing efforts, such as LLM-CXR~\cite{lee2023llm}, often employ parameter sharing and joint multi-task heads for integrated learning. This approach, unfortunately, can introduce task competition and feature interference, which degrades performance in both understanding and generation. HealthGPT~\cite{lin2025healthgpt} mitigates this issue through task-specific H-LoRA modules, offering a structured compromise but not a fundamental solution. Moreover, most current unified medical foundation models still rely on discretized generation paradigms, whose outputs are constrained by vocabulary granularity and fail to recover fine structural details in medical images. A straightforward alternative is to attach a diffusion model to a pre-trained vision–language model~\cite{dong2023dreamllm, ge2024seed}. While this improves generative quality to some extent, it fails to fully exploit understanding features to guide generation, thereby underutilizing the potential of a unified architecture.

Through systematic analysis, we identify two intrinsic limitations in existing unified medical foundation models. First, understanding and generation possess conflicting objectives. Understanding requires semantic abstraction, whereas generation demands pixel-level reconstruction. Jointly learning these opposing goals in a shared feature space causes interference. Second, a paradigm mismatch exists between discrete autoregression and continuous imaging. Discrete methods inherently struggle to capture the fine-grained structural details of medical images. Consequently, prior works often resort to superficial stacking or cascading to combine these tasks. This strategy achieves unification only in form and fails to exploit deep architectural synergy between the two capabilities.

To address these challenges, we propose \model{}. This framework fundamentally resolves the tension between semantic processing and visual synthesis. We adopt a \textbf{decoupled dual-branch architecture} to eliminate the understanding-generation conflict. An autoregressive branch focuses on semantic abstraction, while a separate branch handles pixel-level reconstruction. To bridge the paradigm mismatch, the generation branch leverages \textbf{diffusion models}. This design captures the continuous nature of medical images and avoids the granularity loss inherent in discrete tokenization. Finally, we introduce a \textbf{cross-modal self-attention mechanism} to ensure architectural synergy. Unlike superficial stacking, this module dynamically injects understanding features into the diffusion process. This effectively links semantic reasoning with high-fidelity generation.

To further improve data quality and training efficiency, we implement a rigorous data cleaning pipeline and adopt a stagewise optimization strategy. In the first stage, we freeze the generation branch and train only the understanding branch to acquire medical image interpretation capabilities. In the second stage, we freeze the understanding branch and pre-train the generation branch to learn basic image generation. In the third stage, we continue to freeze the understanding branch and fine-tune the generation branch for high-resolution image generation. Thanks to our architectural design and optimization strategy, \model{} achieves dual-task modeling with significantly fewer parameters, maintaining strong vision-language understanding while attaining high-quality generation performance.

The main contributions of this paper are summarized as follows:
\begin{itemize}
\item[$\bullet$] We propose \model{}, a next-generation unified medical foundation model that structurally decouples yet coordinates understanding and generation. To our knowledge, it is among the first efforts to integrate autoregressive and diffusion paradigms in the medical imaging field.
\item[$\bullet$] We introduce a cross-modal self-attention mechanism to bridge the understanding and generation branches. This module seamlessly integrates the understanding features as contextual conditions, providing dynamic, content-aware guidance throughout the generation process.
\item[$\bullet$] Experiments on chest X-ray report generation and image synthesis tasks show that \model{} uses only a quarter of the parameters of LLM-CXR, yet improves understanding performance (Micro-F1) by 46.1\% and generation performance (FD-Raddino) by 24.2\%, achieving performance comparable to single-task medical foundation models.
\end{itemize}

\section{Related Work}

\begin{figure*}[t]
  \centering
  \includegraphics[width=1.0 \linewidth]{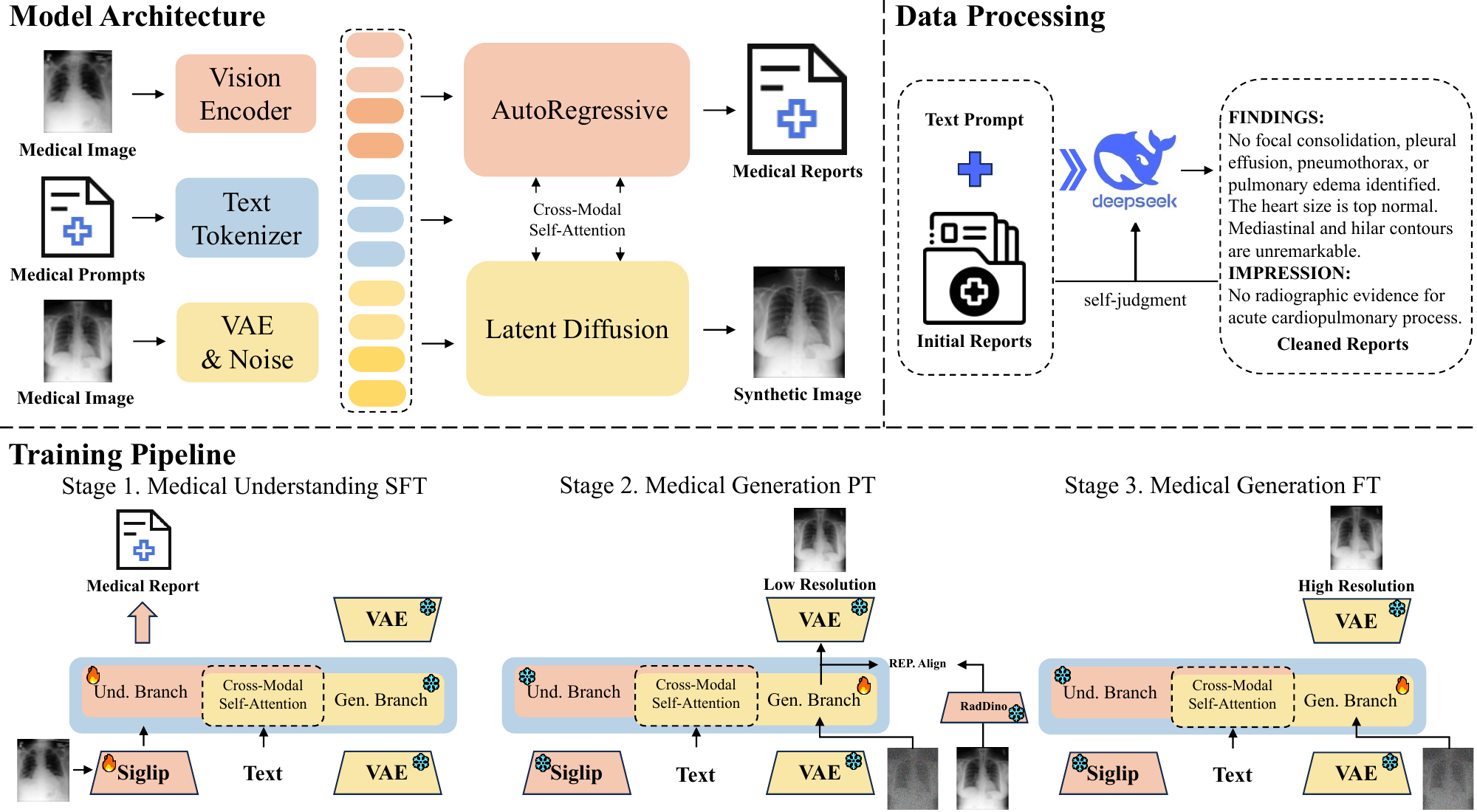}
  \caption{\textbf{Model Architecture.} \model{} comprises two decoupled yet synergistic branches: an autoregressive understanding branch for semantic encoding, and a diffusion-based generation branch for visual synthesis. To enable effective collaboration between them, we introduce a cross-modal self-attention mechanism that allows semantic features to dynamically guide the generation process. \textbf{Data Processing and Training Pipeline.} To fully exploit the potential of this architecture, we design a rigorous data cleaning pipeline and a three-stage training strategy.  This strategy progressively freezes the branches during different stages, ensuring efficient knowledge transfer and stable training.}
  \label{fig:TecPic}
\vspace{-3mm} 
\end{figure*}

\subsection{Single-task Medical Foundation Model}
Medical foundation models for single tasks primarily center on two major objectives. The first category focuses on image understanding. These models handle tasks such as disease diagnosis, knowledge-based question answering, and report generation. Early studies~\cite{yao2024eva, cui2024lkcell, ma2025fully} mainly targeted disease classification. They employed CNNs~\cite{lecun2002gradient} or Transformers~\cite{vaswani2017attention} to extract imaging features, followed by task-specific heads for classification, segmentation, or prediction.

Recently, with the rise of multimodal large language models~\cite{yang2023dawn, hurst2024gpt, wang2024qwen2, li2024llava}, medical foundation models~\cite{tu2024towards, li2023llava, chaves2024towards, tanno2025collaboration, bai2025bench} have evolved toward more clinically meaningful applications, such as medical question answering and report generation. These models typically combine a visual encoder with a large language model, processing multimodal tokens in an autoregressive way.

The second category centers on image generation. These models address tasks like synthesis, super-resolution, and inpainting. With recent advances in generative AI~\cite{rombach2022high, lipman2022flow, yao2024fasterdit, xu2025pixel, yao2025reconstruction, zou2025turbo, deng2025emerging}, diffusion-based approaches have become the dominant paradigm for high-fidelity medical image generation. Single-task generative models can produce realistic, instruction-aligned images, helping to expand datasets and mitigate long-tail issues. Some studies~\cite{dutt2025chexgenbench, bluethgen2025vision} further demonstrate that synthetic data can enhance the performance of visual understanding models, revealing a potential synergy between generation and understanding.

\subsection{Unified Medical Foundation Model}
Unified models~\cite{ge2024seed, xie2024show, chen2025janus, wu2024vila, li2025synergen, wu2024liquid} aim to handle both understanding and generation within a single architecture. They represent a promising next step for medical foundation models. Existing unified medical foundation models~\cite{lee2023llm, zhang2025medunifier, kim2023unixgen} typically adopt a shared Transformer backbone with multi-task heads. However, this parameter-sharing strategy faces inherent conflicts. Understanding tasks require compressing and abstracting information, while generative tasks demand preserving and reconstructing details. These opposing objectives cause feature interference and limit overall performance.

To address this, HealthGPT~\cite{lin2025healthgpt} introduces H-LoRA modules to separate task-specific parameters. This improves performance but still lags behind specialized single-task models. In addition, most unified medical foundation models rely on discrete generation methods based on visual bag-of-words~\cite{van2017neural, esser2021taming}. Since these approaches compress continuous pixel data into a fixed codebook, they inevitably discard high-frequency details and subtle texture variations. Consequently, they struggle to capture continuous, fine-grained pathological patterns. As a result, image fidelity remains limited.

In contrast, \model{} introduces a dual-branch architecture that integrates autoregressive and diffusion paradigms, echoing insights from BAGEL~\cite{deng2025emerging} on the value of bottleneck-free multimodal interaction. This design resolves the objective conflict through architectural decoupling. The understanding branch focuses on semantic comprehension, while the diffusion branch specializes in high-fidelity image generation. The two branches interact via cross-modal self-attention, allowing understanding features to guide the generation process dynamically. Through this synergy, \model{} achieves strong performance comparable to single-task models while maintaining high parameter efficiency.

\section{Method}
In this section, we introduce \model{}, a next-generation medical foundation model designed to achieve decoupled yet synergistic learning between Chest X-ray understanding and generation.

As shown in Figure~\ref{fig:TecPic}, our model contains two core components: an autoregressive understanding branch and a diffusion-based generation branch. The understanding branch, built on a vision-language model, handles semantic abstraction and report reasoning. The generation branch is built directly upon the inherited LLM backbone from the understanding branch, and it specializes in synthesizing high-fidelity Chest X-ray images. A cross-modal self-attention module connects the two, allowing dynamic feature exchange and semantic conditioning during generation.

\subsection{Understanding via Autoregression}
The understanding branch formulates multimodal comprehension as an autoregressive sequence modeling problem. This formulation aligns naturally with medical report generation, where the model must reason over both visual and textual contexts in a causal manner.

Concretely, we define a multimodal token sequence $S = [V,\ T_{in},\ T_{out}]$, where $V$, $T_{in}$, and $T_{out}$ denote the visual tokens, input textual tokens, and output textual tokens respectively. Let $m$ be the starting index of $T_{out}$ in $S$, and $n$ be the ending index of $S$. The cross-entropy loss is computed over the autoregressive predictions for all tokens in $T_{out}$:

\begin{equation}
\mathcal{L}_{CE}=-\sum_{i=m}^{n-1}{\log p\left( S_{i+1}|S_{\le i};\omega_{u} \right)},
\end{equation}
where $n$ is the index of the last token in the sequence $S$, and $\omega_u$ denotes parameters of the understanding branch. This design allows the model to jointly capture visual semantics and linguistic reasoning within a unified space.

\subsection{Generation via Latent Diffusion}
The generation branch adopts a latent diffusion framework that reconstructs medical images from high-level semantics extracted by the understanding branch. Instead of operating in pixel space, diffusion is performed in a VAE-encoded latent space, which greatly improves efficiency and stability.

Given a latent variable $x_t$ sampled from the noisy distribution $p_t(x)$ at time $t$, the model learns to estimate the target velocity field $u_t(x)$ by minimizing the mean-square error:
\begin{equation}
\mathcal{L}_{MSE}=\mathbb{E}_{t,p_t\left( x \right)}\left[ \lVert v_t\left( x;\omega_{g} \right) -u_t\left( x \right) \rVert ^2 \right],
\end{equation}
where $\omega_g$ represents the parameters of the generation branch. The semantic embeddings from the understanding branch act as conditioning inputs, enabling disease-specific synthesis and improved lesion localization.

\subsection{Cross-Modal Self-Attention}
To enable semantically informed visual generation, we introduce a cross-modal self-attention mechanism~\cite{deng2025emerging} that facilitates bidirectional information flow between the understanding and generation branches. Unlike conventional cross-attention, which conditions one modality on a static context, our formulation performs joint self-attention over a unified multimodal token sequence. This design allows semantic representations from the understanding branch to directly modulate the generative trajectory, while also permitting generative states to feed back into the semantic space.

Let the unified sequence be $S=\left[ T_{in},\ N\right]$, where \(T_{in}\) denotes the textual tokens produced by the understanding branch and
\(N\) denotes the noise-conditioned latent embeddings from the generation branch. For each token \(S_i\), we compute modality-specific projections for queries, keys, and values as:
\begin{equation}
\{Q_i, K_i, V_i\}
=
\delta^{u}(i)\, W^{u}_{\{q,k,v\}} S_i
+
\delta^{g}(i)\, W^{g}_{\{q,k,v\}} S_i,
\end{equation}
where the modality selectors \(\delta^u(i)\) and \(\delta^g(i)\) are defined as:
\begin{equation}
\delta^u(i)=
\begin{cases}
1, & S_i \in T_{in}, \\
0, & S_i \in N,
\end{cases}
\qquad
\delta^g(i)=1-\delta^u(i).
\end{equation}

This formulation yields two distinct parameter spaces for understanding and generation tokens, while maintaining a shared attention operation across the unified sequence. The resulting attention map is computed in standard form:
\begin{equation}
\mathrm{Attn}(S) = \mathrm{softmax}\!\left( \frac{QK^\top}{\sqrt{d}} \right) V,
\end{equation}
but all cross-modal interactions are learned implicitly through the joint attention scores rather than through explicit conditioning.

This mechanism synchronizes an autoregressive branch for understanding and a diffusion branch for high-fidelity generation, ultimately improving the fidelity and clinical consistency of the generated images.
\subsection{Three-Stage Training Pipeline}
As shown in Figure~\ref{fig:TecPic}, we adopt a three-stage training strategy to progressively align the understanding and generation branches.

\begin{itemize}
\item[$\bullet$] \textbf{Stage 1: Medical Understanding Supervised Fine-Tuning.}  
In this stage, the generation branch is frozen. We fine-tune the visual encoder, visual connector, and language model backbone in the understanding branch using paired medical images and reports. This step helps the model learn the semantic correspondence between images and text. As a result, the understanding branch gains strong abilities in medical image interpretation and report generation. It also serves as a high-level semantic feature provider for the generation branch in later stages.

\item[$\bullet$] \textbf{Stage 2: Medical Generation Pretraining.}  
Here, we freeze the understanding branch and pre-train the generation branch on text–low-resolution image pairs. To accelerate convergence, we apply Representation Alignment~\cite{yu2024representation}, aligning the eighth-layer hidden states of the generation branch’s language model with RadDino image features using a similarity objective. This design enables the generation branch to better utilize high-level semantics from the understanding branch for low-resolution medical image synthesis.

\item[$\bullet$] \textbf{Stage 3: Medical Generation Fine-Tuning.}  
We maintain the same freezing strategy as in Stage~2 and fine-tune the generation branch using text–high-resolution image pairs. During this stage, we extend the positional encoding of the generation branch and remove feature-level supervision. After fine-tuning, the generation branch can synthesize high-resolution medical images with improved report–image alignment, clearer lesion depiction, and higher visual fidelity.
\end{itemize}

\begin{table}[t]
\centering
\footnotesize
\renewcommand{\arraystretch}{1.1}
\begin{tabular*}{\linewidth}{@{\extracolsep{\fill}} c| c c c}
\toprule
\textbf{Hyper Parameters} & \textbf{Stage One} & \textbf{Stage Two} & \textbf{Stage Three} \\
\midrule
Learning rate        & 1e-4                 & 2e-4                 & 1e-4        \\
LR scheduler         & Constant             & Constant             & Constant    \\
Resolution           & 384                  & 256                  & 512         \\
Use REPA             & --                   & True                 & False       \\
REPA loss weight     & --                   & 0.5                  & --          \\
Batch Size           & 256                  & 256                  & 256         \\
Weight decay         & 0.0                  & 0.0                  & 0.0         \\
Gradient norm clip   & 1.0                  & 1.0                  & 1.0         \\
Optimizer            & \multicolumn{3}{c}{\shortstack{AdamW(0.9,\ 0.95,\ $1\mathrm{e}{-15}$)}} \\
Warm-up steps        & 80                   & 2K                   & 0           \\
Training steps       & 3840                 & 75K                 & 5K         \\
\bottomrule
\end{tabular*}
\caption{\textbf{Hyper-parameter settings for three training stages.} Note that we have introduced weights for the multiple loss functions to make them more accessible. Specifically, the ``REPA loss weight'' refers to the weight ratio between the MSE loss and the REPA loss.}
\label{tab:hyperparams}
\end{table}

\begin{table*}[t]
\centering
\footnotesize
\setlength{\tabcolsep}{4.4pt}
\renewcommand{\arraystretch}{1.1}
\begin{tabular}{l c| c c c c| c c c c}
\toprule
\multirow{2}{*}{\textbf{Model}} & \multirow{2}{*}{\textbf{Und. Params}} &
\multicolumn{4}{c|}{\textbf{CheXbert — ``uncertain'' as negative}$\uparrow$} &
\multicolumn{4}{c}{\textbf{CheXbert — ``uncertain'' as positive}$\uparrow$} \\
\cmidrule(lr){3-6} \cmidrule(lr){7-10}
 & & {Micro F1-14} & {Micro F1-5} & {Macro F1-14} & {Macro F1-5} &
       {Micro F1-14} & {Micro F1-5} & {Macro F1-14} & {Macro F1-5} \\
\midrule

\multicolumn{2}{c|}{} & \multicolumn{8}{c}{Signal-task Medical Foundation Model} \\
\midrule

\addlinespace[2pt]
GPT-4V      & --    & 35.5 & 25.8 & 20.4 & 19.6 & 35.6 & 33.3 & 25.3 & 29.6 \\
Med-PaLM M  & 84B   & 53.6 & \textbf{57.9} & \textbf{39.8} & \textbf{51.6} & {--} & {--} & {--} & {--} \\
LLaVA-Rad   & 7B    & \textbf{57.3} & 57.4 & 39.5 & 47.7 & \textbf{57.3} & \textbf{60.2} & \textbf{44.0} & \textbf{53.3} \\
LLaVA-Med   & 7B    & 27.2 & 22.0 & 15.5 & 16.6 & 27.3 & 24.4 & 18.7 & 20.5 \\
FlamingoCXR & 3B    & {--} & {--} & {--} & {--} & 51.9 & 58.0 & {--} & {--} \\
PromptMRG   & $<$1B & 15.3 & 6.0  & 7.8 & 3.5 & 15.0 & 6.9  & 8.4 & 4.1 \\
RGRG   & $<$1B & 38.9 & 47.2 & 23.7 & 40.8 & 37.4 & 49.0  & 24.4 & 42.7 \\
\midrule

\multicolumn{2}{c|}{} & \multicolumn{8}{c}{Unified Medical Foundation Model} \\
\midrule

\addlinespace[2pt]
LLM-CXR     & 12B  & {--} & {--} & {--} & {--} & 36.0 & {--} & 21.1 & {--} \\
HealthGPT   & 3.8B & 24.2 & 25.1 & 14.6 & 18.5 & 25.5 & 28.2 & 17.5 & 22.6 \\
\textbf{UniX}        & \textbf{1.5B} & \cellcolor{blue!20}\textbf{53.6} & \cellcolor{blue!20}\textbf{56.6} & \cellcolor{blue!20}\textbf{33.2} & \cellcolor{blue!20}\textbf{47.3} & \cellcolor{blue!20}\textbf{52.6} & \cellcolor{blue!20}\textbf{57.9} & \cellcolor{blue!20}\textbf{35.5} & \cellcolor{blue!20}\textbf{49.8} \\
\bottomrule
\end{tabular}
\caption{\textbf{Comparison of various medical foundation models on X-ray understanding tasks.} The data reveals that \model{} achieves a substantial improvement in understanding over the unified medical foundation model. Notably, it delivers performance comparable to a larger, single-task medical foundation model, despite having fewer parameters.}
\label{tab:und_table}
\end{table*}

\begin{figure*}[t]
  \centering
  \includegraphics[width=1.0 \linewidth]{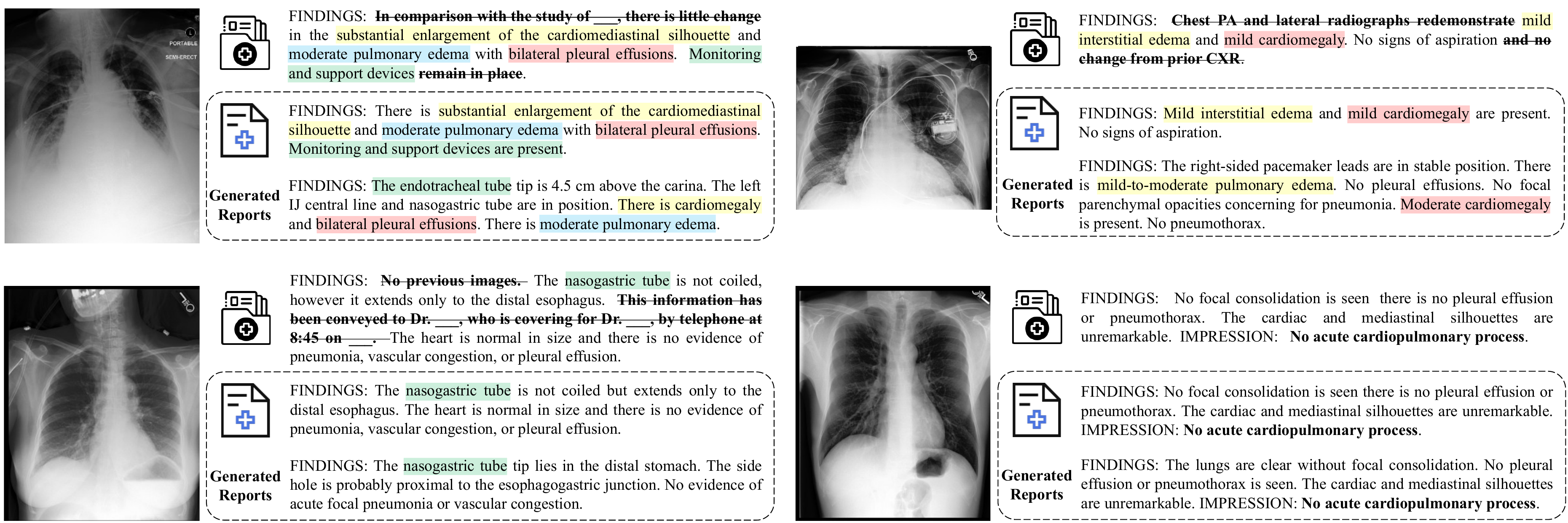}
  \caption{\textbf{Demonstration of Data Processing and Report Generation Efficacy.} The application of large language models enables the purification of raw data by eliminating extraneous information. This process ensures that the model prioritizes and extracts pertinent information related to disease diagnosis.}
  \label{fig:und_visual}
\end{figure*}

\begin{table*}[t]
\centering
\footnotesize 
\setlength{\tabcolsep}{3.1pt}
\renewcommand{\arraystretch}{1.1}
\begin{tabular}{l c c| c c c| c c c c}
\toprule
\textbf{Model} & \textbf{Gen. Params} & \textbf{Resolution} & \textbf{FD-RadDino}$\downarrow$ & \textbf{KD-RadDino}$\downarrow$ & \textbf{Alignment Score}$\uparrow$ & \textbf{Precision}$\uparrow$ & \textbf{Recall}$\uparrow$ & \textbf{Density}$\uparrow$ & \textbf{Coverage}$\uparrow$ \\
\midrule

\multicolumn{3}{c|}{} & \multicolumn{7}{c}{Signal-task Medical Foundation Model} \\
\midrule

Flux.1-Dev $\ast$ & 2.6B & 1024 & 122.400 & 0.144 & 0.036 & 0.420 & 0.008 & 0.125 & 0.326 \\
Lumina 2.0 $\ast$ & 2.5B & 1024 & 101.198 & 0.110 & 0.121 & 0.574 & 0.014 & 0.256 & 0.170 \\
SD V3.5 Medium $\ast$ & 2.5B & 1024 & 91.302 & 0.103 & 0.044 & 0.632 & 0.205 & 0.401 & 0.244 \\
SD V2-1 $\ast$ & 0.86B & 512 & 186.530 & 0.413 & 0.197 & 0.530 & 0.049 & 0.180 & 0.038 \\
RadEdit & 0.86B & 512 & 69.695 & 0.033 & 0.677 & 0.397 & 0.544 & 0.150 & 0.285 \\
Sana $\ast$ & 0.6B & 512 & \textbf{54.225} & \textbf{0.016} & 0.695 & \textbf{0.674} & \textbf{0.614} & \textbf{0.520} & \textbf{0.548} \\
Pixart Sigma $\ast$ & 0.6B & 512 & 60.154 & 0.023 & \textbf{0.697} & 0.666 & 0.522 & 0.506 & 0.506 \\

\midrule

\multicolumn{3}{c|}{} & \multicolumn{7}{c}{Unified Medical Foundation Model} \\
\midrule

LLM-CXR & 12B & 256 & 71.243 & 0.061 & 0.319 & \textbf{0.782} & 0.041 & \textbf{0.671} & 0.459 \\
\textbf{UniX} & \textbf{1.5B} & 256 & \cellcolor{blue!20}65.208 & \cellcolor{blue!20}0.051 & \cellcolor{blue!20}0.251 & \cellcolor{blue!20}0.675 & \cellcolor{blue!20}0.243 & \cellcolor{blue!20}0.366 & \cellcolor{blue!20}0.419 \\
\textbf{UniX} & \textbf{1.5B} & 512 & \cellcolor{blue!20}\textbf{54.022} & \cellcolor{blue!20}\textbf{0.024} & \cellcolor{blue!20}\textbf{0.635} & \cellcolor{blue!20}0.736 & \cellcolor{blue!20}\textbf{0.479} & \cellcolor{blue!20}0.536 & \cellcolor{blue!20}\textbf{0.550} \\
\bottomrule
\end{tabular}
\caption{\textbf{Comparison of various medical foundation models on X-ray generation tasks.} Under a standardized benchmark, \model{} matches the output quality of single-task medical models. Furthermore, it demonstrates exceptional performance in both accuracy and diversity. HealthGPT was not included in this test due to the lack of publicly available text-to-image generation code. $\ast$ indicates that all these generative models from the natural image domain were fine-tuned on the same X-ray dataset.}
\label{tab:gen_table}
\end{table*}

\begin{figure*}[t]
  \centering
  \includegraphics[width=1.0\linewidth]{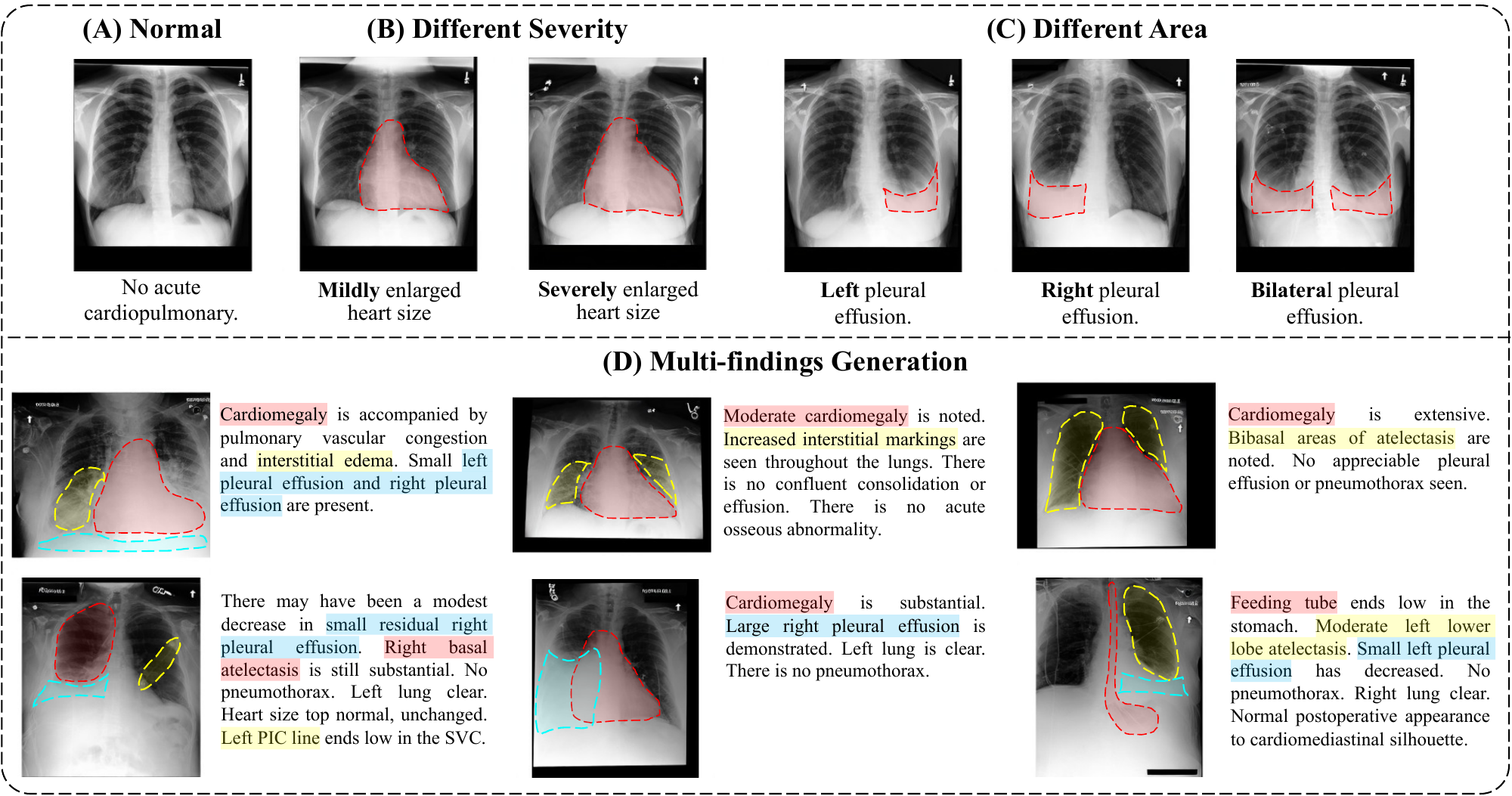}
  \caption{\textbf{Qualitative Examples from \model{}.} \textbf{(A)-(C)} illustrate the model's precise control over the attributes of generated findings, including their severity and location. In \textbf{(D)}, the model successfully synthesizes a complex radiographic scene containing multiple findings that are consistent with a full clinical report, highlighting its ability to process and integrate extensive contextual information.}
  \label{fig:gen_visual}
\end{figure*}

\begin{table*}[t]
\centering
\footnotesize
\setlength{\tabcolsep}{2.55pt}
\renewcommand{\arraystretch}{1.1}
\begin{tabular}{l c c | *{14}{c}}
\toprule
\multirow{2}{*}{\textbf{Model}} & \multirow{2}{*}{\textbf{Gen. Params}} & \multirow{2}{*}{\textbf{Resolution}} &
\multicolumn{14}{c}{\textbf{FD-RadDino$\downarrow$}} \\
\cmidrule(lr){4-17}
& & & At & Cd & Cn & Ec & Fc & Fr & LL & LO & NF & PE & PO & PN & PT & SD \\

\midrule
\multicolumn{3}{c|}{} & \multicolumn{14}{c}{Signal-task Medical Foundation Model} \\
\midrule
RadEdit & 0.86B & 512 & 63.38 & 62.79 & 136.59 & 76.94 & 155.97 & 197.58 & 184.11 & 61.90 & 67.88 & 60.60 & 215.92 & 114.66 & 151.34 & 53.10 \\
Pixart Sigma $\ast$& 0.6B & 512 & 59.27 & 60.39 & 133.96 & 73.93 & 155.53 & 179.44 & 174.63 & 56.83 & 48.74 & 59.05 & 210.90 & 108.42 & 150.55 & 51.61 \\
Sana $\ast$& 0.6B & 512 & \textbf{51.03} & \textbf{54.68} & \textbf{127.46} & \textbf{67.84} & \textbf{147.00} & \textbf{172.32} & \textbf{163.14} & \textbf{49.23} & \textbf{44.60} & \textbf{49.80} & \textbf{199.45} & \textbf{88.52} & \textbf{141.99} & \textbf{46.51} \\

\midrule
\multicolumn{3}{c|}{} & \multicolumn{14}{c}{Unified Medical Foundation Model} \\
\midrule

LLM-CXR & 12B & 256 & 71.57 & 71.37 & 136.65 & 83.18 & 148.28 & \textbf{168.50} & 163.22 & 66.93 & 64.62 & 67.83 & 200.84 & 108.04 & 147.52 & 67.54 \\
\textbf{UniX} & \textbf{1.5B} & 256 & \cellcolor{blue!20}63.34 & \cellcolor{blue!20}63.39 & \cellcolor{blue!20}129.32 & \cellcolor{blue!20}73.88 & \cellcolor{blue!20}150.25 & \cellcolor{blue!20}177.68 & \cellcolor{blue!20}165.88 & \cellcolor{blue!20}58.31 & \cellcolor{blue!20}60.58 & \cellcolor{blue!20}58.55 & \cellcolor{blue!20}201.53 & \cellcolor{blue!20}105.96 & \cellcolor{blue!20}141.63 & \cellcolor{blue!20}57.61 \\
\textbf{UniX} & \textbf{1.5B} & 512 & \cellcolor{blue!20}\textbf{52.19} & \cellcolor{blue!20}\textbf{51.70} & \cellcolor{blue!20}\textbf{122.84} & \cellcolor{blue!20}\textbf{64.36} & \cellcolor{blue!20}\textbf{142.23} & \cellcolor{blue!20}176.35 & \cellcolor{blue!20}\textbf{156.81} & \cellcolor{blue!20}\textbf{49.15} & \cellcolor{blue!20}\textbf{45.71} & \cellcolor{blue!20}\textbf{48.06} & \cellcolor{blue!20}\textbf{191.65 }& \cellcolor{blue!20}\textbf{99.31} & \cellcolor{blue!20}\textbf{135.48} & \cellcolor{blue!20}\textbf{47.04} \\
\bottomrule
\end{tabular}
\caption{\textbf{Generation Performance per Pathology.} Within the unified medical foundation model, \model{} dominates the comparison, achieving top performance in 13 out of the 14 categories. $\ast$ indicates that all these generative models from the natural image domain were fine-tuned on the same X-ray dataset.}
\label{tab:pathology}
\end{table*}

\section{Experiments}
This section, we describe how \model{} exploits its decoupled autoregressive–diffusion dual-branch design through data processing, model configuration, and a three-stage training pipeline.

\subsection{Implementation Details}
\begin{itemize}
\item[$\bullet$] \textbf{Data Details.} We conduct experiments on the MIMIC-CXR dataset~\cite{johnson2019mimic}.
For the understanding branch, we use frontal-view radiographs and refine the paired reports with the DeepSeek large language model. The full cleaning pipeline is provided in the supplementary material. Following the official split, we obtain 163,344 image–report pairs for training and 2,365 for testing.
For the generation branch, we follow the processing and split protocol of ChexGenBench~\cite{dutt2025chexgenbench}, resulting in 237,387 training and 4,352 test pairs. Images are resized to 384×384 for understanding fine-tuning, 256×256 for generation pre-training, and 512×512 for generation fine-tuning. Aspect ratios are preserved by padding when needed.
\item[$\bullet$] \textbf{Model Details.} Both branches are partially initialized from Janus-Pro~\cite{chen2025janus}.
For the understanding branch, we adopt \texttt{siglip-large-patch16-384}~\cite{zhai2023sigmoid} as the visual encoder. It produces 1024-dimensional embeddings, which are mapped to the 2048-dimensional LLM space through a two-layer MLP. The language backbone contains 24 transformer layers and incorporates QK normalization and QKV bias.
For the generation branch, we use a 16× downsampled, 16-channel VAE for encoding and decoding. Two single-layer MLPs provide bidirectional projection between the 16-dimensional latent space and the 2048-dimensional language space. The generation backbone follows the same initialization strategy as the understanding branch.
\item[$\bullet$] \textbf{Training Details.} All models are trained with full-parameter fine-tuning on eight NVIDIA L20 GPUs. The complete hyperparameter configuration is listed in Table~\ref{tab:hyperparams}.
\end{itemize}

\subsection{Decoupled Architecture for Task Separation}
We conduct extensive comparisons among understanding-only, generation-only, and unified medical foundation models. Qualitative examples of \model{}’s performance in both understanding and generation are shown in Figures~\ref{fig:und_visual} and~\ref{fig:gen_visual}. As observed, \model{} produces precise reports and high-fidelity images. We next present detailed results for understanding and generation tasks.

For understanding tasks, we primarily evaluate the medical reliability of generated reports using the CheXbert F1 score~\cite{smit2020chexbert} between generated and ground-truth reports. Additional evaluation metrics, including BLEU~\cite{papineni2002bleu}, Radgraph~\cite{jain2021radgraph} and ROUGE-L~\cite{lin2004rouge}, are provided in the supplementary material. For generation tasks, we measure generation quality with FD-RadDino and KD-RadDino, assess image–text consistency with the Alignment Score, and evaluate accuracy and diversity using the four PRDC metrics.

\subsubsection{Understanding}
The training dynamics of the understanding branch are detailed in the Supplementary Material. As training progresses, the cross-entropy loss decreases steadily, and the model’s report generation ability improves rapidly at first before gradually plateauing. Although continued fine-tuning can further reduce the loss, we observed that key performance metrics begin to decline, indicating that the model tends to overfit specific patterns rather than learning general principles.

Table~\ref{tab:und_table} summarizes the understanding performance of various medical foundation models. \model{} achieves better performance than unified models while using substantially fewer parameters. Compared with single-task models, it outperforms models of similar scale and approaches the performance of much larger ones.
We exclude medical agent systems here since they typically build upon existing medical foundation models and rely on multi-model collaboration. A comparison between \model{} and such agents is provided in the supplementary material.

\subsubsection{Generation}
We observed that the mean squared error for the generation branch decreases rapidly early on and then slows in both stages. During medical generation pre-training, key metrics show limited improvement between 25K and 50K steps but increase sharply from 50K to 75K steps. The corresponding loss curves and a detailed analysis are included in the Supplementary Material.

Table~\ref{tab:gen_table} reports the generation results. Compared with LLM-CXR, \model{} delivers clear improvements in both image quality and image–text alignment. It also consistently outperforms single-task models, reflecting the advantage of its decoupled yet collaborative architecture. Notably, \model{} performs on par with the strong baseline Sana. Even though Sana is also fine-tuned on the target dataset, \model{} matches its performance and achieves comparable, and in some metrics slightly superior. This demonstrates that our unified approach maintains top-tier generation quality without compromising semantic consistency.

We further evaluate pathology-specific generation quality in Table~\ref{tab:pathology}. \model{} achieves consistently higher fidelity across a wide range of lesion types, capturing subtle pathological cues and preserving clinically relevant details. In these fine-grained tasks, \model{} remains highly competitive with Sana. This is particularly noteworthy given that \model{} balances a unified multi-task objective whereas Sana focuses on a specialized generation task. Consequently, the results demonstrate strong fine-grained visual synthesis capability and robust performance under diverse diagnostic conditions.

\section{Ablations}
\subsection{Impact of Data Cleaning on Understanding}
Figure~\ref{fig:und_visual} illustrates the critical role of data cleaning with DeepSeek in mitigating hallucinations. Raw hospital reports often contain significant noise, such as underscores, technical metadata, and conversational fillers, which complicates the alignment between visual features and textual descriptions. By employing targeted prompts to strip away these non-diagnostic elements, we construct a cleaner and more semantic-dense target for the model. This preprocessing step is crucial because it forces the model to attend strictly to clinically relevant patterns during training, resulting in generated reports that are factually grounded and free from structural hallucinations.

\subsection{Impact of Joint Dual-Branch Optimization}
\begin{table}[t]
\centering
\footnotesize
\setlength{\tabcolsep}{3.2pt}
\begin{tabular*}{\linewidth}{@{\extracolsep{\fill}} c c| c c c c}
\toprule
\multicolumn{2}{c|}{\textbf{Configs}} 
    & \multicolumn{4}{c}{\textbf{Metrics}} \\
\cmidrule(r){1-2} \cmidrule(l){3-6}
\textbf{Train Branch} & \textbf{Data Ratio}
    & \textbf{Micro-F1}$\uparrow$ & \textbf{Macro-F1}$\uparrow$ & \textbf{FD}$\downarrow$ & \textbf{KD}$\downarrow$ \\
\midrule
Gen        & 0 : 1   & 53.2 & 36.0 & 62.114 & 0.041 \\
Und \& Gen & 1 : 1   & 43.7 & 29.5 & 65.747 & 0.047 \\
Und \& Gen & 1 : 2   & 42.9 & 28.3 & 76.125 & 0.064 \\
Und \& Gen & 1 : 4   & 44.9 & 28.4 & 76.108 & 0.064 \\
Und \& Gen & 0 : 1   & 13.9 & 6.3  & 74.815 & 0.054 \\
\bottomrule
\end{tabular*}
\caption{\textbf{Impact of Joint Dual-Branch Optimization.} Freezing the understanding branch yields fast generative gains without harming comprehension. Unfreezing it without understanding data severely degrades comprehension and offers no generative benefit. Mixing both data types mitigates this degradation but slows generative learning.}
\label{tab:ablation_configs}
\end{table}

To study how joint fine-tuning affects understanding and generation, we design multiple medical image generation experiments, each fine-tuned for 2K generation steps. We evaluate three strategies:
\begin{itemize}
\item[1)] Unfreeze only the generation branch.
\item[2)] Unfreeze both branches and train with a mixture of understanding and generation data, where we vary the mixing ratio.
\item[3)] Unfreeze both branches but train without any understanding data.
\end{itemize}

The results in Table~\ref{tab:ablation_configs} lead to several observations. First, the fine-tuned understanding branch should not be involved in subsequent training of the generation branch. Fully freezing the understanding branch yields rapid gains in generation performance without harming understanding accuracy. In contrast, unfreezing the branch without providing understanding data is detrimental; it severely degrades understanding performance and offers no benefit to generation. Adjusting the ratio of understanding and generation data can partially mitigate the drop in understanding performance, as the semantic supervision stabilizes the updated parameters. However, this setup forces the model to balance two competing objectives within the same updates, which slows the acquisition of strong generative capability.

\section{Conclusion}
We present UniX, a next-generation unified medical foundation model that achieves architectural decoupling and coordination for Chest X-Ray understanding and generation. Existing unified medical foundation models neither resolve the intrinsic conflict nor fully exploit the strengths of different modeling paradigms.
To address these limitations, we design a dual-branch architecture that combines autoregressive understanding with diffusion-based generation. This structure decouples the two tasks and prevents mutual interference. We introduce a cross-modal self-attention mechanism that aligns both branches and enables understanding features to guide generation dynamically.
With these designs, UniX delivers stronger understanding and generation performance than prior unified medical foundation models while using fewer parameters, and it reaches competitiveness with dedicated single-task medical foundation models. We hope that \model{} offers a new perspective for advancing medical foundation models.

{
    \small
    \bibliographystyle{ieeenat_fullname}
    \bibliography{main}
}
\end{document}